\documentclass{article}
\usepackage{spconf,amsmath,graphicx}
\usepackage{cite}
\usepackage{siunitx}
\usepackage{amsmath,amssymb,amsfonts}
\usepackage{graphicx}
\usepackage{textcomp}
\usepackage{booktabs}
\usepackage{xcolor}
\usepackage{bm}
\usepackage{esvect}
\usepackage[hidelinks]{hyperref} 
\def\BibTeX{{\rm B\kern-.05em{\sc i\kern-.025em b}\kern-.08em
    T\kern-.1667em\lower.7ex\hbox{E}\kern-.125emX}}
\usepackage{cite}
\usepackage{url}
\usepackage{amsmath,amssymb,amsfonts}
\usepackage{algorithm}
\usepackage{graphicx}
\usepackage{changepage}
\usepackage{algpseudocode}

\usepackage{lipsum}
\usepackage{ctable}
\usepackage{color, colortbl}

\usepackage{array, booktabs, makecell, multirow}
\definecolor{Gray}{gray}{0.9}
\usepackage{textcomp}
 
\usepackage{textcomp}
\usepackage{booktabs, makecell, multirow, tabularx}
\setcellgapes{2pt}

\usepackage{stfloats}

\usepackage{lipsum}
\usepackage{booktabs}
\newcommand\mytab[1]{\begin{tabular}[t]{@{}c@{}} #1 \end{tabular}}
\newcommand\mc[2]{\multicolumn{#1}{c}{#2}}

\usepackage{multicol}
\usepackage{subcaption}

\begin{document}
%

\title{HAROOD: Human Activity Classification and Out-of-Distribution Detection with short-range FMCW Radar}
\name{Sabri Mustafa Kahya$^{\star}$ \qquad Muhammet Sami Yavuz$^{\star \dagger}$ \qquad Eckehard Steinbach$^{\star}$}
\address{  $^{\star}$Technical University of Munich,
School of Computation, Information and Technology,\\  Department of Computer Engineering, Chair of Media Technology,\\ Munich Institute of Robotics and Machine Intelligence (MIRMI) \\ $^{\dagger}$Infineon Technologies AG}


\maketitle

%
\begin{abstract}
We propose HAROOD as a short-range FMCW radar-based human activity classifier and out-of-distribution (OOD) detector. It aims to classify human sitting, standing, and walking activities and to detect any other moving or stationary object as OOD. We introduce a two-stage network. The first stage is trained with a novel loss function that includes intermediate reconstruction loss, intermediate contrastive loss, and triplet loss. The second stage uses the first stage's output as its input and is trained with cross-entropy loss. It creates a simple classifier that performs the activity classification. On our dataset collected by 60 \si{\GHz} short-range FMCW radar, we achieve an average classification accuracy of 96.51\%. Also, we achieve an average AUROC of 95.04\% as an OOD detector. Additionally, our extensive evaluations demonstrate the superiority of HAROOD over the state-of-the-art OOD detection methods in terms of standard OOD detection metrics.

\end{abstract}
\section{Introduction}
\label{sec:intro}

In recent years, the affordability of short-range radars has increased their popularity in both industry and academia. Also, thanks to their robustness to environmental conditions like lighting, rain, and smoke and their privacy-preserving nature, they are preferred for applications such as human heartbeat estimation, gesture recognition, presence detection, and people counting\cite{vital_sign,b33,kahya2023hood,people_counting}. Here, we address the human activity classification application by also considering the out-of-distribution (OOD) samples.

OOD detection has gained huge attention in the last decade\cite{b5,b27,b25}. It aims to prevent the overconfident predictions of the samples that lie outside of the training domain. Therefore, it is very crucial, especially for safety-critical applications. In addition to its robust human activity classification capability, HAROOD also serves as an OOD detector to prevent the classification of the OOD samples to the classes seen during training.

This work proposes a novel framework that can accurately classify human sitting, standing, and walking activities and detect the OOD samples. Our system depends on a novel architecture trained in two stages. The first stage provides a novel loss function incorporating intermediate reconstruction loss mainly for OOD detection, intermediate contrastive loss for outlier exposure (OE) that yields better OOD detection, and triplet loss for better classification performance. The second stage of our training uses the embeddings from the triplet training and performs the final classification of human activities. Our key contributions are as follows:
\vspace{-0.2cm}
\begin{itemize}
\item We propose a novel pipeline that accurately and simultaneously performs human activity classification and OOD detection tasks. The first stage of the architecture consists of multi-encoder multi-decoder pairs followed by a set of convolutional and linear layers that provide the final embeddings. The second stage performs like a simple and lightweight convolutional classifier that takes the embeddings as its input.
\vspace{-0.25cm}
\item We propose a novel loss function consisting of intermediate reconstruction loss, intermediate contrastive loss, and triplet loss. The contrastive training part performs as a novel OE technique that pushes away the OOD samples from the in-distribution (ID) samples, yielding better OOD detection performance.
\vspace{-0.25cm}
\item We perform extensive experiments and achieve an average human activity classification accuracy of 96.51\% and an average AUROC of 95.04\% for OOD detection. Also, we show that HAROOD outperforms the state-of-the-art (SOTA) OOD detection methods in terms of common OOD detection metrics and test time.

\end{itemize}

\section{Related Work}
\label{sec:related}

 In the radar-based human activity classification field, a pioneering study employed a Support Vector Machine (SVM) classifier, utilizing a collection of specially crafted features derived from the micro-Doppler spectrogram \cite{firstHAC}. \cite{b34} combined the Unscented Kalman filter and LSTM for better classification. \cite{bi-lstm} proposed a Bidirectional LSTM model to classify six different activities. DL-based solutions \cite{paramConv,DNNAct} also exist in the literature. \cite{eucDistance} introduced a DL network with a novel Euclidean distance-based softmax layer to learn activity labels from Doppler spectrograms.  Additional studies leveraged DNNs based on micro-Doppler images for classification \cite{dnnBasedMicro, dnnBasedMicro2}. Some studies \cite{dnnWithOOD, dnnWithOOD2, dataDrivenAct} have incorporated additional non-human activity classes, such as waving curtains and empty room scenarios, to mitigate misclassification into human activity categories. However, these works are robust only to those anomalies used also during training. Our work proposes a global (anomaly-type free) OOD detection solution in addition to its robust activity classification capability.

The significance of OOD detection was first highlighted by a seminal study \cite{b1} that utilized maximum softmax probabilities. It claims that OODs have lower softmax scores than IDs. By applying input perturbation and temperature scaling on logits, ODIN \cite{b2} aimed to increase the softmax scores of IDs.  \cite{b3} applied model ensembling approach on top of \cite{b2}. G-ODIN \cite{b30} built upon \cite{b2} with a new training scheme. MAHA \cite{b4} uses intermediate representations and provides a Mahalanobis distance-based OOD detector. FSSD \cite{b6} also benefits from intermediate feature representations for detection. \cite{b31} uses non-parametric KNN applied on the embeddings of the penultimate layer for detection. With the help of the $logsumexp$ function, \cite{b7} provided an energy-based OOD detection method. ReAct\cite{b28} applies truncation to unit activations in the penultimate layer and can be used together with a variety of OOD detection methods. GradNorm \cite{b14} uses gradients' vector norm, which is backpropagated from the KL divergence between the softmax output and a uniform probability distribution. MaxLogit \cite{hendrycks2022scaling}  utilizes maximum logit scores to differentiate between IDs and OODs. \cite{b8} introduced the OE technique by using a limited number of OODs during training to push their softmax scores toward a uniform distribution. OECC \cite{b10} demonstrated improvement for OE with a novel loss, including additional regularization terms. In this work, we also provide a novel OE strategy for improved OOD detection.

In the radar domain, there are also some OOD detection studies. Using 60 \si{\GHz} FMCW radar, \cite{RB-OOD} aims to classify any moving object other than a walking person as OOD. MCROOD \cite{MCROOD} is a multi-class radar OOD detector that targets to identify disturbers apart from some human activities. With its multi-encoder multi-decoder network, \cite{kahya2023hood} simultaneously addresses the human presence and OOD detection problems. Here, we also address human activity classification and OOD detection applications at the same time.

\section{Radar Configuration \& Pre-processing}

Our radar sensor has one transmit (Tx) antenna and three receiver (Rx) antennas. The Tx antenna emits $N_c$ chirp signals, and the Rx antennas receive the reflected signals. The transmitted and received signals are mixed and low-pass filtered, yielding the intermediate frequency (IF) signal. Then, an analog-to-digital converter (ADC) is applied to obtain raw ADC data. The data is organized as $N_c \times N_s$, where $N_c$ and $N_s$ correspond to slow-time and fast-time samples, respectively. With three Rx antennas ($N_{Rx}$), the digitized signal shape becomes $N_{Rx} \times N_c \times N_s$.

We use both macro and micro range-Doppler images (RDIs) as input in our architecture. To generate the macro RDI, we apply \textbf{Range-FFT} to the fast-time signal for range information. Mean removal is applied to have single-channel range information followed by the moving target identification (\textbf{MTI}) operation, which eliminates static targets. Then, \textbf{Doppler-FFT} on the slow-time signal captures phase variations and produces the macro RDI. For micro RDI, we again apply \textbf{Range-FFT} for range information. By stacking eight range spectrograms and applying mean removal to both fast and slow times, we effectively reduce noise. The application of \textbf{Sinc} filtering enhances target signals. Finally, performing \textbf{Doppler-FFT} along the slow-time dimension results in the generation of the micro RDI. To enhance the capture of human body movements, the last step in the preprocessing pipeline for both RDIs involves applying \textbf{E-RESPD} \cite{kahya2023hood}.

\begin{figure*}[htbp]
\centerline{\includegraphics[width=0.81\linewidth]{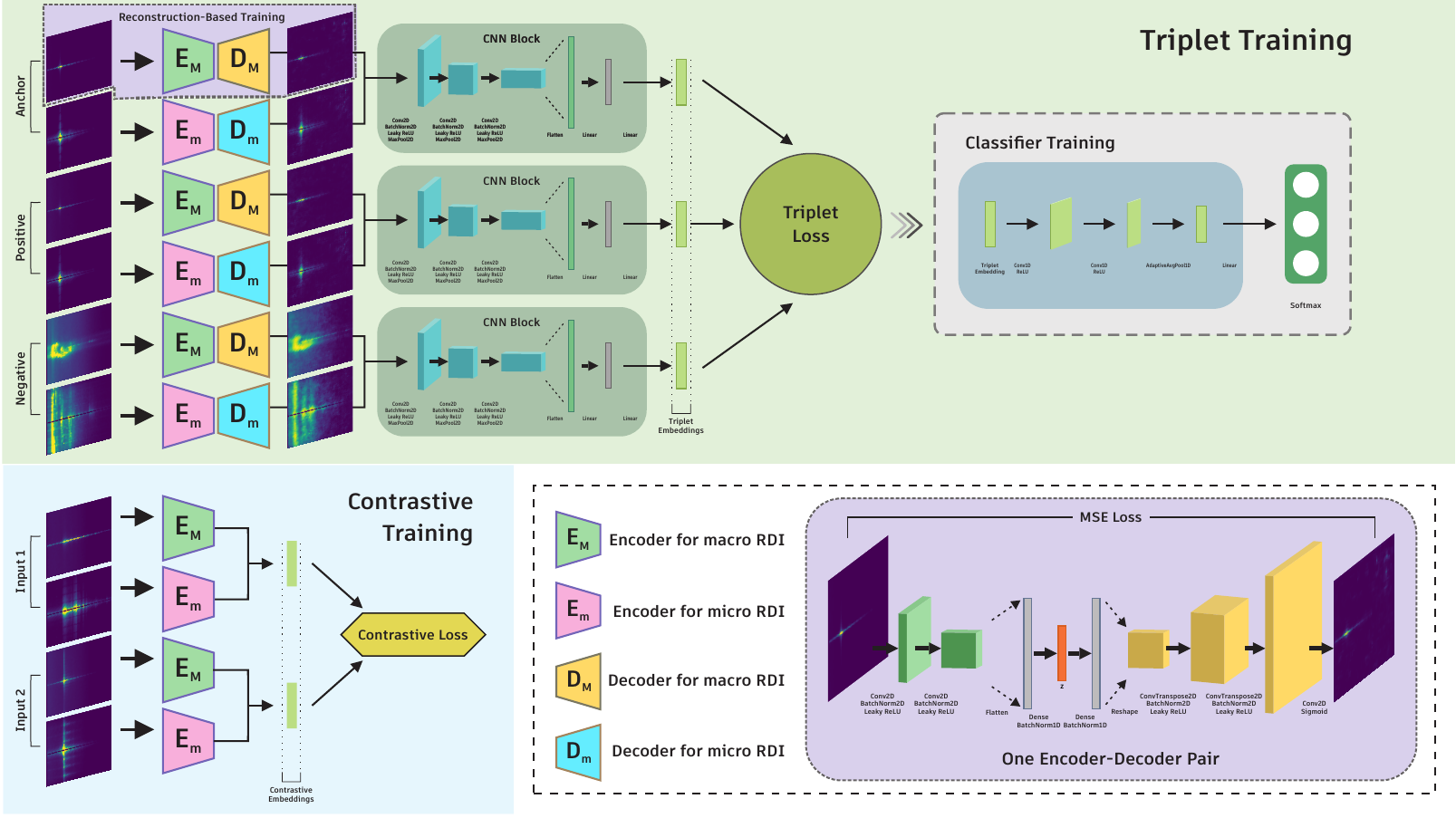}}
\caption{\footnotesize Detailed training scheme of HAROOD. The same blocks have shared weights. Here, we have two stages of training. The first stage involves reconstruction-based, triplet, and contrastive training. The encoder-decoder pairs perform the reconstruction training and aim for accurate OOD detection. The CNN block following the Encoder-Decoder pairs utilizes the reconstructed outputs to finalize triplet training and generate the embeddings. The contrastive training performs the OE operation by exposing limited OOD data to the network. It helps to have clearly separable ID and OOD sets. The first stage performs simultaneous training. The second stage is classifier training. It trains a simple classifier to perform precise human activity classification using the embeddings. A zoom-in version of one E-D pair is also provided with a purple background.}
\label{fig:pipline}
\end{figure*}

\section{Problem Statement and HAROOD}
HAROOD concurrently tackles both OOD detection as a binary classification task and human activity recognition as a multi-class classification challenge.
Our novel architecture comprises two stages (See Figure \ref{fig:pipline}). \textbf{In the first stage}, we have two encoder-decoder pairs followed by convolutional layers. The encoder-decoder pairs are separately designed for macro and micro RDIs. The following block creates the final embeddings. In this stage, we use our novel loss function, including intermediate reconstruction, intermediate contrastive and triplet loss, and Adamax optimizer (a variant of the Adam optimizer that relies on the infinity norm).
\textbf{The encoder-decoder} pairs are specialized for OOD detection purposes. Since the triplet training part takes inputs from the reconstruction part, the encoder-decoder pairs are also trained with triples. We use mean-squared-error (MSE) loss for each encoder-decoder pair. Since we use triples and have two pairs, we have six MSEs, which we call intermediate reconstruction loss:
\vspace{-0.1cm}
\begin{equation*}
\begin{aligned}
\mathcal{L}_{rec}  = \frac{1}{b} \sum_{k \in \{a,p,n\}} \sum_{j \in \{m,M\}} \sum_{i=1}^{b} (\textbf{X}^{(i)}_{k,j} - D_{j}(E_{j}(\textbf{X}^{(i)}_{k,j})))^2,
\end{aligned}
\label{eq:loss_recons}
\
\end{equation*}
where ${X}^{(i)}$ is input, $b$ is the batch size, and $k$ is an index for the sample types of anchor (a), positive (p), and negative (n). $j$ indicates macro or micro RDIs; $E_j$ and $D_j$ correspond to the encoder and decoder for the RDI type $j$. 

\textbf{The triplet training} takes its input (anchor, positive, negative) from the reconstructed output of the decoders. The reconstructed macro and micro RDIs are first concatenated and fed to a convolutional block that produces the embeddings of the triples. This part aims to create an easily distinguishable embedding space for each activity class by minimizing the triplet loss:
\begin{equation*}
\vspace{-0.1cm}
\begin{aligned}
\mathcal{L}_{tri}&  = \frac{1}{b} \sum_{i=1}^{b} \max(\|\textbf{e}^{(i)}_{t,a} -  \textbf{e}^{(i)}_{t,p}\|_2 - \| \textbf{e}^{(i)}_{t,a} - \textbf{e}^{(i)}_{t,n}\|_2 + \alpha_t, 0),
\end{aligned}
\label{eq:loss_triplet}
\
\end{equation*}
where $\textbf{{e}}^{(i)}_{t,a},\textbf{{e}}^{(i)}_{t,p}$, and $\textbf{{e}}^{(i)}_{t,n}$ are the triplet embeddings for the anchor (a), positive (p), and negative (n) instances. $b$ is the batch size, and $\alpha_t$ is the margin set to 2. 

\textbf{The contrastive training} part is designed for OE purposes and only trains the encoders. We also use limited OOD samples. It assumes ID classes as one class and OODs as the other class.  The contrastive training part aims to decrease the reconstruction quality of the OOD data by pushing their embeddings from the ID embeddings only in three epochs:
\begin{equation*}
\vspace{-0.1cm}
\begin{aligned}
\mathcal{L}_{con}  = \frac{1}{b} \sum_{i=1}^{b} &(1-Y)\|\textbf{e}^{(i)}_{c,1} -  \textbf{e}^{(i)}_{c,2}\|^2_2 +  \\
&Y(\max(0, \alpha_c - \|\textbf{e}^{(i)}_{c,1} -  \textbf{e}^{(i)}_{c,2}\|_2))^2,          
\end{aligned}
\label{eq:loss_contrast}
\
\end{equation*}
where $b$ is the batch size, $\textbf{e}^{(i)}_{c,1}$ ($\textbf{e}^{(i)}_{c,2}$) is the contrastive embedding that is the concatenation of encoded macro and micro RDIs. $\alpha_c$ is the margin, which is 2, and $Y$ equals 0 if the samples in a pair are similar, 1 otherwise.

\begin{figure}[htbp]
\centerline{\includegraphics[width=0.90\columnwidth]{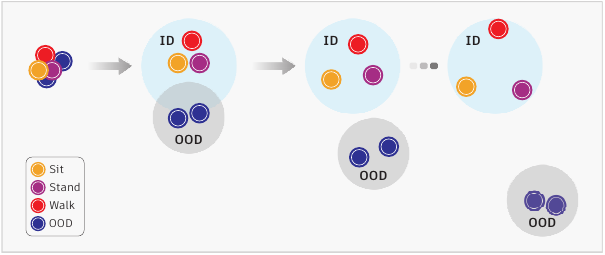}}
\caption{\small  The ultimate goal of our training scheme.}
\label{fig:train-goal}
\vspace{-0.25cm}
\end{figure}

\textbf{The total loss function} becomes $\mathcal{L}  =\mathcal{L}_{rec} + \mathcal{L}_{tri} + \mathcal{L}_{con}$. Our network is a fast learner, so we train it in just six epochs. (three epochs with $\mathcal{L}_{rec}, \mathcal{L}_{tri}, \mathcal{L}_{con}$ three epochs with $\mathcal{L}_{rec}, \mathcal{L}_{tri}$). Figure \ref{fig:train-goal} explains the ultimate aim of our novel training scheme. The ID and OOD sets are separated, while the ID embeddings are pushed from each other within their own set.
\textbf{The OOD detection} is performed in the reconstruction. Since encoder-decoder pairs are trained only with IDs, and we perform OE, we expect that the reconstruction MSEs of OODs are greater than those of IDs. We calculate the reconstruction scores from the weighted summation of the MSEs from both encoder-decoder pairs ($1$ for macro $0.001$ for micro). A simple threshold is defined in a way that 95\% (changes based on the application) of ID data is correctly detected. In the end, if the total error of a sample is more than this threshold, it is OOD, ID otherwise.
\textbf{The second stage} is a simple classifier that takes the 1D embedding as input and classifies the input as either sit, stand, or walk. It is trained with cross-entropy loss and Adam optimizer.

\section{Experiments}
We conduct our experiments utilizing an NVIDIA GeForce RTX 3070 GPU, an Intel Core i7-11800H CPU, and a 32GB DDR4 RAM module.
\begin{table*}[ht]
\centering
\footnotesize
\caption{\small OOD Detection Results. Performance comparison with other popular methods. All values are shown in percentages. $\uparrow$ indicates that higher values are better, while $\downarrow$ indicates that lower values are better.}
\label{Results}
\setlength\tabcolsep{0pt}
\begin{tabular*}{\textwidth}{@{\extracolsep{\fill}}c|ccccccccccccc}
    \toprule
     
    & \mc{4}{\mytab{Sit}} & \mc{4}{\mytab{Stand}}
    & \mc{4}{\mytab{Walk}} &{Test Time }\\
  
    \cmidrule{2-5} \cmidrule{6-9} \cmidrule{10-13}
\centering
     Methods 
    & AUROC
    & AUPR\textsubscript{IN}
    & AUPR\textsubscript{OUT}
    & FPR95 
    & AUROC 
    & AUPR\textsubscript{IN}
    & AUPR\textsubscript{OUT}
    & FPR95 
    & AUROC 
    & AUPR\textsubscript{IN}
    & AUPR\textsubscript{OUT}
    & FPR95 
    & (seconds) \\
    &  $\uparrow$
    &  $\uparrow$
    &  $\uparrow$
    &  $\downarrow$
    &  $\uparrow$
    &  $\uparrow$
    &  $\uparrow$
    &  $\downarrow$
    &  $\uparrow$
    &  $\uparrow$
    &  $\uparrow$
    &  $\downarrow$
    &  $\downarrow$ \\  
      
    \midrule
     ODIN \cite{b2} &49.58&25.03&78.08&80.12&64.43&40.57&79.58&92.17&47.09&31.87&65.24&91.70&401 \\

     MSP \cite{b1} &67.22&38.02&86.86&60.01&19.74&17.24&56.00&100&45.56&30.07&73.05&69.18&91   \\

     ENERGY \cite{b7} &46.43&32.19&66.92&99.15&76.62&65.91&87.10&84.66&38.35&29.15&64.42&84.78&91   \\

    MAHA \cite{b4} &77.30&67.75&85.56&87.07&59.67&44.42&77.84&89.06&74.94&55.31&84.95&69.61&1480 \\
    FSSD \cite{b6} &14.45&17.08&58.64&93.15&79.18&76.79&88.17&82.14&\textbf{98.63}&\textbf{97.70}&\textbf{99.24}&\textbf{9.38}&1278\\
    OE\cite{b8} &41.90&23.14&66.25&99.73&62.43&47.40&76.43&95.41&81.32&82.13&78.71&100&91  \\
    GRADNORM\cite{b14} &90.09&\textbf{74.27}&95.33&45.00&24.02&19.06&57.08&100&13.61&21.99&47.20&99.71&282  \\        
    REACT\cite{b28} &71.52&57.82&83.10&87.73&22.41&18.42&56.39&100&50.06&34.30&64.36&95.63&97 \\
    MAXLOGIT\cite{hendrycks2022scaling} &53.94&35.90&72.88&95.03&38.13&21.51&69.12&94.93&38.14&27.85&60.99&94.95&97  \\
    
    \midrule
    \midrule
    \addlinespace

HAROOD&\textbf{93.57}&72.81&\textbf{97.83}&\textbf{16.51}&\textbf{95.43}&\textbf{89.04}&\textbf{97.95}&\textbf{14.29}&96.13&94.22&97.76&23.21&\textbf{28} \\

    
    \bottomrule

\end{tabular*}
\end{table*}
We use Infineon’s BGT60TR13C 60 \si{\GHz} FMCW radar sensor for data collection. We record the movements of four individuals and commonly seen objects in indoor settings across various rooms, including houses, offices, and school rooms. Among these rooms, 13 are only used for training, while 8 are only used for inference. The radar is positioned at 2.5 meters with a 30-degree tilt to the ground. The data includes both ID and OOD samples. IDs are human walking, sitting, and standing activities, with varying distances of 1 to 4 meters from the radar sensor. OODs consist of various household moving objects such as table and stand fans, a remote-controlled (RC) toy car, a vacuum cleaner, a robot vacuum cleaner, swinging laundry, blinds, curtains, boiling water from a kettle, running water from a tap, and empty room with stationary disturbers. In our balanced training set, we use a total of 227467 ID frames. For OE, we use 21222 ID and 6234 OOD (only from two types: table fan and RC toy car) frames. We use 61304 ID frames and 47409 OOD frames in the balanced test set.

For OOD detection, we use four common metrics. \textbf{AUROC} measures the area under the receiver operating characteristic (ROC) curve. \textbf{AUPR\textsubscript{IN/OUT}} refers to the area under the precision-recall curve when considering ID/OOD samples as positives. \textbf{FPR95} represents the false positive rate (FPR) when the true positive rate (TPR) reaches 95\%.  For human activity classification, we use \textbf{accuracy} as an evaluation metric. We also present \textbf{Test Time}, indicating the inference time in seconds required to evaluate all test samples.

\begin{table}[ht]
\footnotesize
\caption{\small Human Activity Classification Results. }
\centering
\begin{tabular}{@{\extracolsep{\fill}}ccccc}
\toprule   
{} &{}&\multicolumn{1}{c}{Accuracy } &&  {Average Accuracy}  \\

 \cmidrule{2-4} 
\centering 
  Methods & Sit  & Stand & Walk\\ 
\midrule
RESNET \cite{resnet} &91.38&89.91&99.63&94.38\\
 HAROOD    &\textbf{92.43}&\textbf{95.95}&\textbf{99.83}&\textbf{96.51}   \\

\bottomrule
\end{tabular}
\label{table:ActClass} 
\end{table}

\begin{figure}[ht]

    \centering
    \begin{subfigure}[b]{0.5\columnwidth}
    \includegraphics[width=\textwidth]{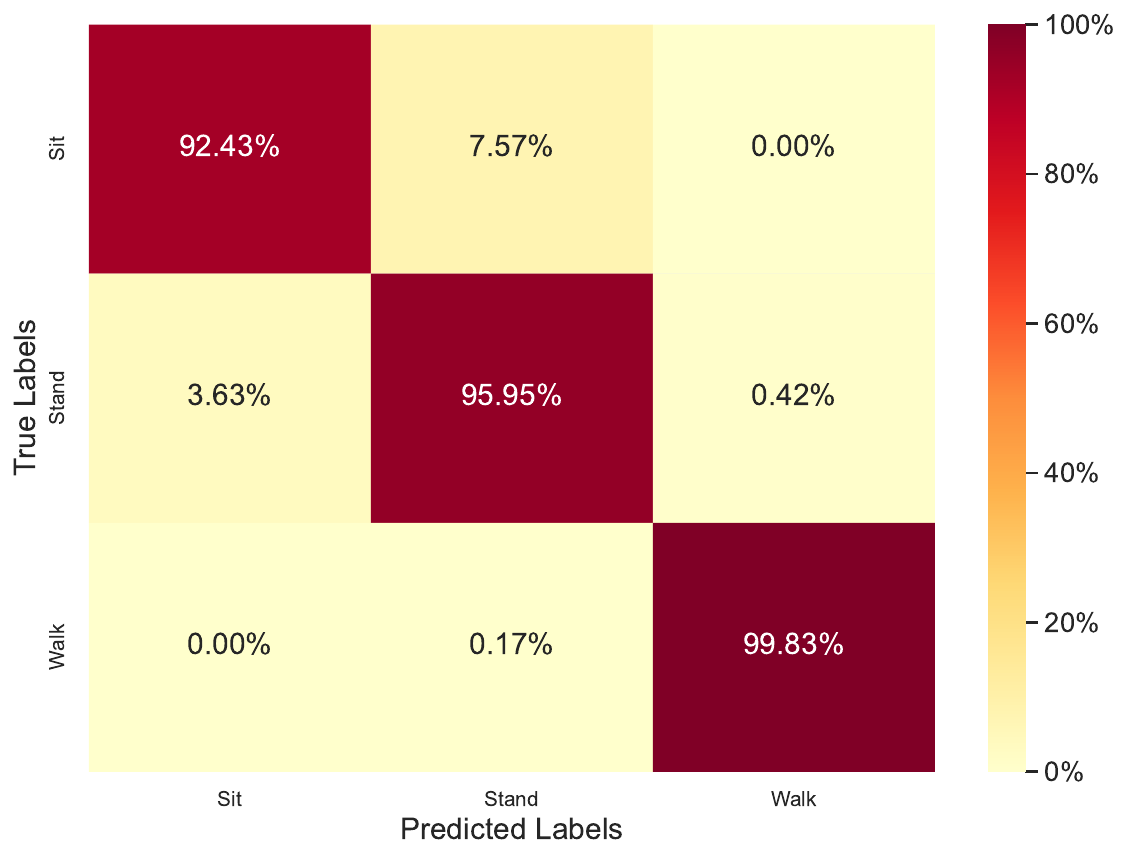}
    \caption{HAROOD}
    \label{fig:conf_matrix_HAROOD}
    \end{subfigure}%
    \begin{subfigure}[b]{0.5\columnwidth}
    \includegraphics[width=\textwidth]{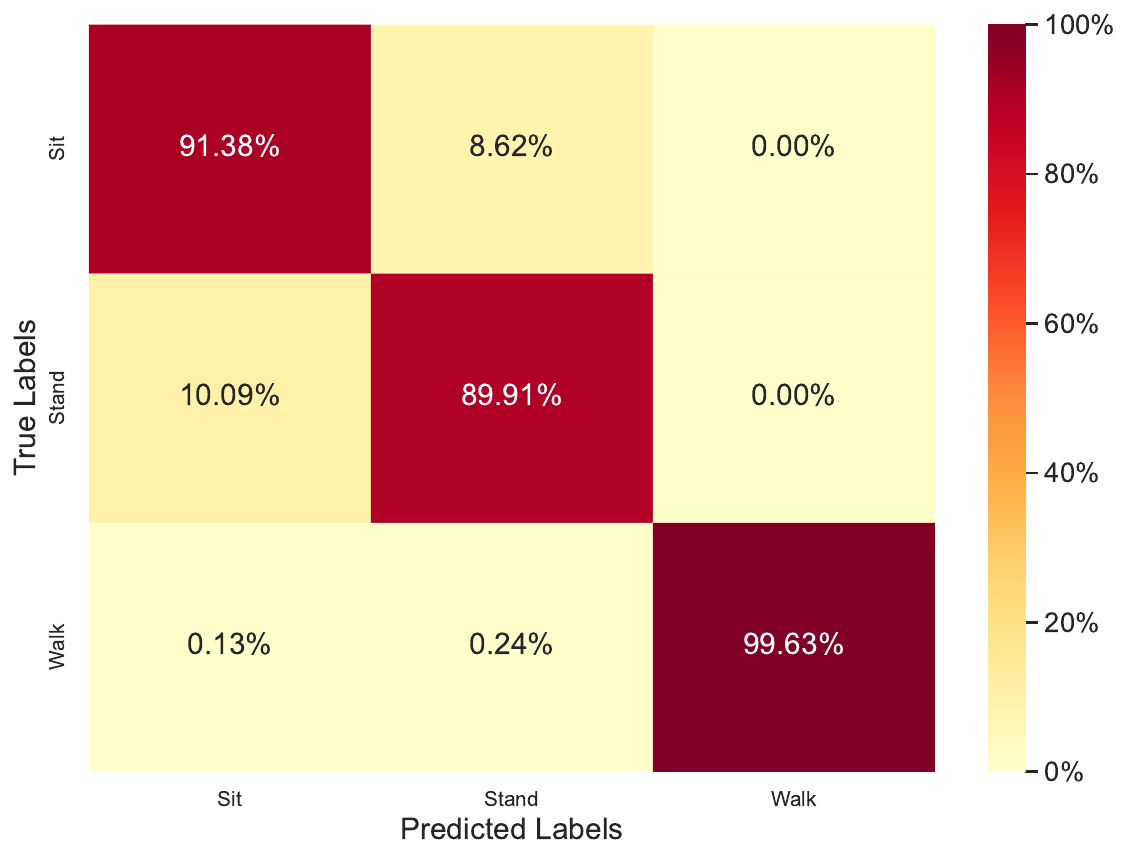}
    \caption{ResNet}
    \label{fig:conf_matrix_resnet}
    \end{subfigure}%
     \caption{\small Confusion matrices for our method HAROOD (a) and ResNet model (b). }
     \label{fig:conf_matrices}
\end{figure}

We compare HAROOD against nine SOTA methods using common OOD metrics. We train a ResNet34 \cite{resnet} backbone with our ID data by treating it as a three-class multi-class classification problem. The pre-trained model is then utilized to apply the SOTA methods, and the performance of each method is evaluated as in Table \ref{Results}. Our model's classification performance is also presented in Table \ref{table:ActClass}, including a comparison with the ResNet. Figure \ref{fig:conf_matrices} provides the confusion matrices for both HAROOD and ResNet\cite{resnet}.
\vspace{0.205cm}
\section{Conclusion}
This paper presents HAROOD, a novel short-range FMCW radar-based human activity classifier and OOD detector. HAROOD utilizes a two-stage network architecture with a unique loss function and a novel OE technique in its first stage. It achieves an average classification accuracy of 96.51\% for human sitting, standing, and walking activities and an average AUROC of 95.04\% for OOD detection on our 60 \si{\GHz} short-range FMCW radar dataset. Comparisons with SOTA methods further confirm the superior performance of HAROOD. 

\vfill \pagebreak


\bibliographystyle{ieeetr}

\footnotesize

\begin{thebibliography}{10}

\bibitem{vital_sign}
Muhammad Arsalan, Avik Santra, and Christoph Will,
\newblock ``Improved contactless heartbeat estimation in fmcw radar via kalman filter tracking,''
\newblock {\em IEEE Sensors Letters}, vol. 4, no. 5, pp. 1--4, 2020.

\bibitem{b33}
Souvik Hazra and Avik Santra,
\newblock ``Robust gesture recognition using millimetric-wave radar system,''
\newblock {\em IEEE Sensors Letters}, vol. 2, no. 4, pp. 1--4, 2018.

\bibitem{kahya2023hood}
Sabri~Mustafa Kahya, Muhammet~Sami Yavuz, and Eckehard Steinbach,
\newblock ``Hood: Real-time robust human presence and out-of-distribution detection with low-cost fmcw radar,'' 2023.

\bibitem{people_counting}
Cem~Yusuf Aydogdu, Souvik Hazra, Avik Santra, and Robert Weigel,
\newblock ``Multi-modal cross learning for improved people counting using short-range fmcw radar,''
\newblock in {\em 2020 IEEE International Radar Conference (RADAR)}, 2020, pp. 250--255.

\bibitem{b5}
Chandramouli~Shama Sastry and Sageev Oore,
\newblock ``Detecting out-of-distribution examples with {G}ram matrices,''
\newblock in {\em International Conference on Machine Learning (ICML)}, 2020.

\bibitem{b27}
Jingkang Yang, Haoqi Wang, Litong Feng, Xiaopeng Yan, Huabin Zheng, Wayne Zhang, and Ziwei Liu,
\newblock ``Semantically coherent out-of-distribution detection,''
\newblock in {\em IEEE International Conference on Computer Vision (ICCV)}, 2021.

\bibitem{b25}
Haoqi Wang, Zhizhong Li, Litong Feng, and Wayne Zhang,
\newblock ``Vim: Out-of-distribution with virtual-logit matching,''
\newblock {\em arXiv}, 2022.

\bibitem{firstHAC}
Youngwook Kim and Hao Ling,
\newblock ``Human activity classification based on micro-doppler signatures using a support vector machine,''
\newblock {\em IEEE Transactions on Geoscience and Remote Sensing}, vol. 47, no. 5, pp. 1328--1337, 2009.

\bibitem{b34}
Prachi Vaishnav and Avik Santra,
\newblock ``Continuous human activity classification with unscented kalman filter tracking using fmcw radar,''
\newblock {\em IEEE Sensors Letters}, vol. 4, no. 5, pp. 1--4, 2020.

\bibitem{bi-lstm}
Aman Shrestha, Haobo Li, Julien Le~Kernec, and Francesco Fioranelli,
\newblock ``Continuous human activity classification from fmcw radar with bi-lstm networks,''
\newblock {\em IEEE Sensors Journal}, vol. 20, no. 22, pp. 13607--13619, 2020.

\bibitem{paramConv}
Thomas Stadelmayer, Avik Santra, Robert Weigel, and Fabian Lurz,
\newblock ``Parametric convolutional neural network for radar-based human activity classification using raw adc data,''
\newblock {\em IEEE Sensors J}, vol. 21, no. 17, pp. 19529--19540, 2020.

\bibitem{DNNAct}
Xinyu Li, Yuan He, and Xiaojun Jing,
\newblock ``A deep multi-task network for activity classification and person identification with micro-doppler signatures,''
\newblock in {\em 2019 International Radar Conference (RADAR)}, 2019, pp. 1--5.

\bibitem{eucDistance}
Thomas Stadelmayer, Markus Stadelmayer, Avik Santra, Robert Weigel, and Fabian Lurz,
\newblock ``Human activity classification using mm-wave fmcw radar by improved representation learning,''
\newblock in {\em Proceedings of the 4th ACM Workshop on Millimeter-Wave Networks and Sensing Systems}, New York, NY, USA, 2020, mmNets'20, Association for Computing Machinery.

\bibitem{dnnBasedMicro}
Yuan He, Yang Yang, Yue Lang, Danyang Huang, Xiaojun Jing, and Chunping Hou,
\newblock ``Deep learning based human activity classification in radar micro-doppler image,''
\newblock in {\em 2018 15th European Radar Conference (EuRAD)}, 2018, pp. 230--233.

\bibitem{dnnBasedMicro2}
Youngwook Kim and Taesup Moon,
\newblock ``Human detection and activity classification based on micro-doppler signatures using deep convolutional neural networks,''
\newblock {\em IEEE Geoscience and Remote Sensing Letters}, vol. 13, no. 1, pp. 8--12, 2016.

\bibitem{dnnWithOOD}
Rodrigo Hernangómez, Avik Santra, and Sławomir Stańczak,
\newblock ``Human activity classification with frequency modulated continuous wave radar using deep convolutional neural networks,''
\newblock in {\em 2019 International Radar Conference (RADAR)}, 2019, pp. 1--6.

\bibitem{dnnWithOOD2}
Rodrigo Hernang{\'o}mez, Avik Santra, and S{\l}awomir Sta{\'n}czak,
\newblock ``Study on feature processing schemes for deep-learning-based human activity classification using frequency-modulated continuous-wave radar,''
\newblock {\em IET Radar, Sonar \& Navigation}, vol. 15, no. 8, pp. 932--944, 2021.

\bibitem{dataDrivenAct}
Thomas Stadelmayer, Avik Santra, Robert Weigel, and Fabian Lurz,
\newblock ``Data-driven radar processing using a parametric convolutional neural network for human activity classification,''
\newblock {\em IEEE Sensors Journal}, vol. 21, no. 17, pp. 19529--19540, 2021.

\bibitem{b1}
Dan Hendrycks and Kevin Gimpel,
\newblock ``A baseline for detecting misclassified and out-of-distribution examples in neural networks,''
\newblock in {\em International Conference on Learning Representations (ICLR)}, 2017.

\bibitem{b2}
Shiyu Liang, Yixuan Li, and R.~Srikant,
\newblock ``Enhancing the reliability of out-of-distribution image detection in neural networks,''
\newblock in {\em International Conference on Learning Representations (ICLR)}, 2018.

\bibitem{b3}
Balaji Lakshminarayanan, Alexander Pritzel, and Charles Blundell,
\newblock ``Simple and scalable predictive uncertainty estimation using deep ensembles,''
\newblock in {\em Advances in Neural Information Processing Systems}. 2017, vol.~30, Curran Associates, Inc.

\bibitem{b30}
Y.~C. {Hsu}, Y.~{Shen}, H.~{Jin}, and Z.~{Kira},
\newblock ``Generalized odin: Detecting out-of-distribution image without learning from out-of-distribution data,''
\newblock in {\em IEEE/CVF Conference on Computer Vision and Pattern Recognition (CVPR)}, 2020.

\bibitem{b4}
Kimin Lee, Kibok Lee, Honglak Lee, and Jinwoo Shin,
\newblock ``A simple unified framework for detecting out-of-distribution samples and adversarial attacks,''
\newblock in {\em International Conference on Neural Information Processing Systems (NeurIPS)}, 2018.

\bibitem{b6}
Haiwen Huang, Zhihan Li, Lulu Wang, Sishuo Chen, Bin Dong, and Xinyu Zhou,
\newblock ``Feature space singularity for out-of-distribution detection,''
\newblock in {\em Proceedings of the Workshop on Artificial Intelligence Safety 2021 (SafeAI 2021)}, 2021.

\bibitem{b31}
Yiyou Sun, Yifei Ming, Xiaojin Zhu, and Yixuan Li,
\newblock ``Out-of-distribution detection with deep nearest neighbors,''
\newblock in {\em International Conference on Machine Learning (ICML)}, 2022.

\bibitem{b7}
Weitang Liu, Xiaoyun Wang, John Owens, and Yixuan Li,
\newblock ``Energy-based out-of-distribution detection,''
\newblock in {\em Advances in Neural Information Processing Systems (NeurIPS)}, 2020.

\bibitem{b28}
Yiyou Sun, Chuan Guo, and Yixuan Li,
\newblock ``React: Out-of-distribution detection with rectified activations,''
\newblock in {\em Advances in Neural Information Processing Systems (NeurIPS)}, 2021.

\bibitem{b14}
Rui Huang, Andrew Geng, and Yixuan Li,
\newblock ``On the importance of gradients for detecting distributional shifts in the wild,''
\newblock in {\em Advances in Neural Information Processing Systems (NeurIPS)}, 2021.

\bibitem{hendrycks2022scaling}
Dan Hendrycks, Steven Basart, Mantas Mazeika, Andy Zou, Joe Kwon, Mohammadreza Mostajabi, Jacob Steinhardt, and Dawn Song,
\newblock ``Scaling out-of-distribution detection for real-world settings,''
\newblock {\em International Conference on Machine Learning (ICML)}, 2022.

\bibitem{b8}
Dan Hendrycks, Mantas Mazeika, and Thomas Dietterich,
\newblock ``Deep anomaly detection with outlier exposure,''
\newblock in {\em International Conference on Learning Representations (ICLR)}, 2019.

\bibitem{b10}
Aristotelis-Angelos Papadopoulos, Mohammad~Reza Rajati, Nazim Shaikh, and Jiamian Wang,
\newblock ``Outlier exposure with confidence control for out-of-distribution detection,''
\newblock {\em Neurocomputing}, vol. 441, pp. 138--150, 2021.

\bibitem{RB-OOD}
Sabri~Mustafa Kahya, Muhammet~Sami Yavuz, and Eckehard Steinbach,
\newblock ``Reconstruction-based out-of-distribution detection for short-range fmcw radar,''
\newblock {\em arXiv}, 2023.

\bibitem{MCROOD}
Sabri~Mustafa Kahya, Muhammet Sami~Yavuz, and Eckehard Steinbach,
\newblock ``Mcrood: Multi-class radar out-of-distribution detection,''
\newblock in {\em ICASSP 2023 - 2023 IEEE International Conference on Acoustics, Speech and Signal Processing (ICASSP)}, 2023, pp. 1--5.

\bibitem{resnet}
Kaiming He, Xiangyu Zhang, Shaoqing Ren, and Jian Sun,
\newblock ``Deep residual learning for image recognition,''
\newblock in {\em IEEE Conference on Computer Vision and Pattern Recognition (CVPR)}, 2016.

\end{thebibliography}

\end{document}